%% file: 00_main.tex
\documentclass[letterpaper]{article} 
\usepackage{aaai23}  
\usepackage{times}  
\usepackage{helvet}  
\usepackage{courier}  
\usepackage[hyphens]{url}  
\usepackage{graphicx} 
\def\UrlFont{\rm}  
\usepackage{natbib}  
\usepackage{caption} 
\usepackage{algorithm}
\usepackage{algorithmic}
\usepackage{subfigure}
\usepackage{tikz}
\usepackage{newfloat}
\usepackage{listings}
\usepackage{tabularx}
\newcommand{\egnows}{e.g.}
\newcommand{\eg}{e.g.\ }
\newcommand*\circled[1]{\tikz[baseline=(char.base)]{\node[shape=circle,draw,minimum size=4mm, inner sep=0pt] (char){#1};}}

\DeclareCaptionStyle{ruled}{labelfont=normalfont,labelsep=colon,strut=off} 
\floatstyle{ruled}
\newfloat{listing}{tb}{lst}{}
\floatname{listing}{Listing}
\title{Using ScrutinAI for Visual Inspection of DNN Performance in a Medical Use Case}
\author {
    Rebekka Görge,\textsuperscript{\rm 1}
    Elena Haedecke,\textsuperscript{\rm 2,\rm 1}
    Michael Mock\textsuperscript{\rm 1}
}
\affiliations {
    \textsuperscript{\rm 1}Fraunhofer IAIS, 53757 St. Augustin, Germany\\
    \textsuperscript{\rm 2}University of Bonn, 53113 Bonn, Germany\\
    rebekka.goerge@iais.fraunhofer.de, elena.haedecke@iais.fraunhofer.de, michael.mock@iais.fraunhofer.de
}

\begin{document}
\maketitle

\begin{abstract}
Our Visual Analytics (VA) tool ScrutinAI supports human analysts to investigate interactively model performanceand data sets. 
Model performance depends on labeling quality to a large extent. 
In particular in medical settings, generation of high quality labels requires in depth expert knowledge and is very costly. 
Often, data sets are labeled by collecting opinions of groups of experts.
We use our VA tool to analyse the influence of label variations between different experts on the model performance.
ScrutinAI facilitates to perform a root cause analysis that distinguishes weaknesses of deep neural network (DNN) models caused by varying or missing labeling quality from true weaknesses.
We scrutinize the overall detection of intracranial hemorrhages and the more subtle differentiation between subtypes in a publicly available data set.
\end{abstract}

\input{01_introduction.tex}
\input{02_case_study.tex}
\input{03_conclusion.tex}
\input{04_acknowledgement}
\bibliography{99_references_aaai23}
\end{document}

%% file: 01_introduction.tex
\section{Introduction}
Machine Learning (ML) models, especially based on Deep Learning (DL), promise a high potential for medical applications. 
To use machine based decisions in medical practice, models must be reliable and transparent.
Additionally to local explanations of single model decisions, a detailed understanding of the deep neural network (DNN) is needed to uncover hidden patterns within the inner structure of the algorithm \cite{tjoa2020survey}. 
The performance and reliability of a supervised model is influenced by aleatoric uncertainty resulting, \egnows, from label noise.
This concerns especially the medical domain \cite{karimi2020deep}: Labeling medical data sets, such as image data,  requires resource intensive domain knowledge which is biased due to subjective expert judgement, annotation habits as well as annotator's errors. 
This results in a lack of consistency among different observers, which is defined as inter-observer variability~\cite{liao2019modelling}.
Final annotations are often a consensus from multiple experts, which challenges appropriate aggregation of these labels. 
As a consequence of the negative affection of label noise on model performance, the trustworthiness of quality metrics based on these labels is questionable. 
As handling label noise is still largely unnoticed in medical domain \cite{karimi2020deep}.
we use Visual Analytics (VA) to analyse the aleatoric uncertainty of labels and base the detection of true weaknesses on this.  

Visual Analytics is a multidisciplinary field where interactive systems and tools are being developed that enable the human analyst to engage in a structured reasoning process by providing appropriate visualizations and representations of the data.
An important aspect is to explicitly benefit from the humans' tacit and expert knowledge by supporting the analyst with specific workflows to gain insights and knowledge into the problem domain~\cite{Sacha2014VAKnowledgeGenerationModel}. 
Integrating the human into the analysis process is especially important in case of complex DNN models~\cite{andrienko2022}.
To further facilitate DNN interpretability, transparency methods, \eg Grad-CAM++ \cite{chattopadhay2018grad}, are often employed. 
For this task, the data and its representations must be prepared in such a way that no relevant information is lost or hidden, and at the same time the workflow and integrated widgets provide enough flexibility for deep dive analysis.
For example, CheXplain~\cite{xie2020chexplain} focuses at addressing the specific needs in the healthcare domain for exploring and understanding a detection model for radiographic chest images.

Our own focus lies on assessing potential model vulnerabilities by leveraging the semantic context of the data, specifically object-level description of entities, through the use of detailed descriptive meta data. 
For this task, we developed the VA tool ScrutinAI~\cite{haedecke_ScrutinAI_eurova}, whose functionalities and workflow promote the generation of semantic hypotheses during the exploration in iterative\textbf{ analysis cycles}.
The meta data, as well as precomputed model predictions of the model, are loaded into the tool via an easily exchangeable CSV file.
Additionally, image data is displayed with options for zooming and overlaying.
Various widgets provide options for filtering and querying the data, such as textual queries, an interactive selection of interesting data points, data distribution along categories, or correlation plots.
Although being originally developed for uses cases in the automotive domain, the modular design of \mbox{ScrutinAI} allows a simple adaption to other domains by the integration of customized widgets.

%% file: 02_case_study.tex
\section{Visual Analysis of the Medical Use Case}
In this paper, we demonstrate the visual analysis of the influence of label noise on model performance within ScrutinAI for the use case of DNN detection of intracranial hemorrhages.
We give details on the medical use case, model and data sets and the modularity and extensibility of ScrutinAI. 
We describe our analysis process within ScrutinAI and our findings, in which we uncover inter-observer variability in the general detection of hemorrhages and among all classes in the data sets, in particular with respect to very similar classes. In consecutive analysis cycles, we reveal a negative influence of this label noise on our model's performance.

\subsection{Background of the Medical Use Case}
Intracranial hemorrhage is an urgent and life-threatening emergency requiring rapid medical treatment. 
To determine region and size of a hemorrhage, imaging techniques such as computed tomography (CT) can be used, consisting of individual slices giving a three-dimensional impression of the head.  
Automatic image recognition using DL can assist doctors with quick detection and characterization.
Therefore, we use a DNN of our prior work \cite{goergeThesis} trained on the biggest publicly available multinational and multi-institutional data set of intracranial CT scans provided by the Radiology Society of North America (RSNA) \cite{flanders2020construction}. For each sample, one of 60 experienced radiologists annotated on slice-level the region of the hemorrhage with the corresponding subtype \textit{any}, \textit{epidural}, \textit{intraparenchymal}, \textit{interaventricular}, \textit{subarachnoid} or \textit{subdural}.
We use the labeled data set part with 80\% train and 20\% test split and window setting\footnote {Typically used by radiologists. Corresponds to gray-value mapping, where a specific interval  of the CT range is selected to highlight different intensity ranges.} as preprocessing method. 
If the model predicts a hemorrhage in a region, we generate with Grad-CAM++ a heatmap as local explanation. 

We evaluate our model in addition to the RSNA test split on the commonly used public CQ500 data set \cite{chilamkurthy2018development}. 
After data selection and cleaning, we obtain 490 CT scans, using only those slice series with the lowest sampling rate.
The data set is annotated on CT-level by three independent senior radiologists with eight, twelve, and 20 years of experience, using the same six subtypes of hemorrhages as in the RSNA data set. 
We use the individual annotations as well as the original ground truth derived from the majority vote of the three radiologists.

In addition to the CT-level annotations, we use labels and bounding boxes on slice-level of \cite{reis2020brain}.
The labels are an aggregation of annotations from three different neuroradiologists with six, four, and less than one year of practice. 
The single annotations of the radiologists have not been published. 
The slice-level labels by \citeauthor{reis2020brain} are uniquely matched to the CT-level annotations by the \textit{SOP-Instance UID} and the \textit{Study-Instance UID}. 
In the following, we treat the annotations of \citeauthor{chilamkurthy2018development} as radiologists number one, two and three, and of \citeauthor{reis2020brain} as fourth radiologist.
To make the annotations of \citeauthor{reis2020brain} as well as the model's output comparable to the others, we generalize them to a CT-level by using the maximum value per region and CT. 
Besides generating more knowledge by combining all various annotations, we enrich the meta data, \egnows, by specifying for each CT the proportion of radiologists detecting a specific hemorrhage.

\subsection{Analysis in ScrutinAI and Results}
\begin{figure*}[t]
    \centering
    \includegraphics[width=\textwidth]{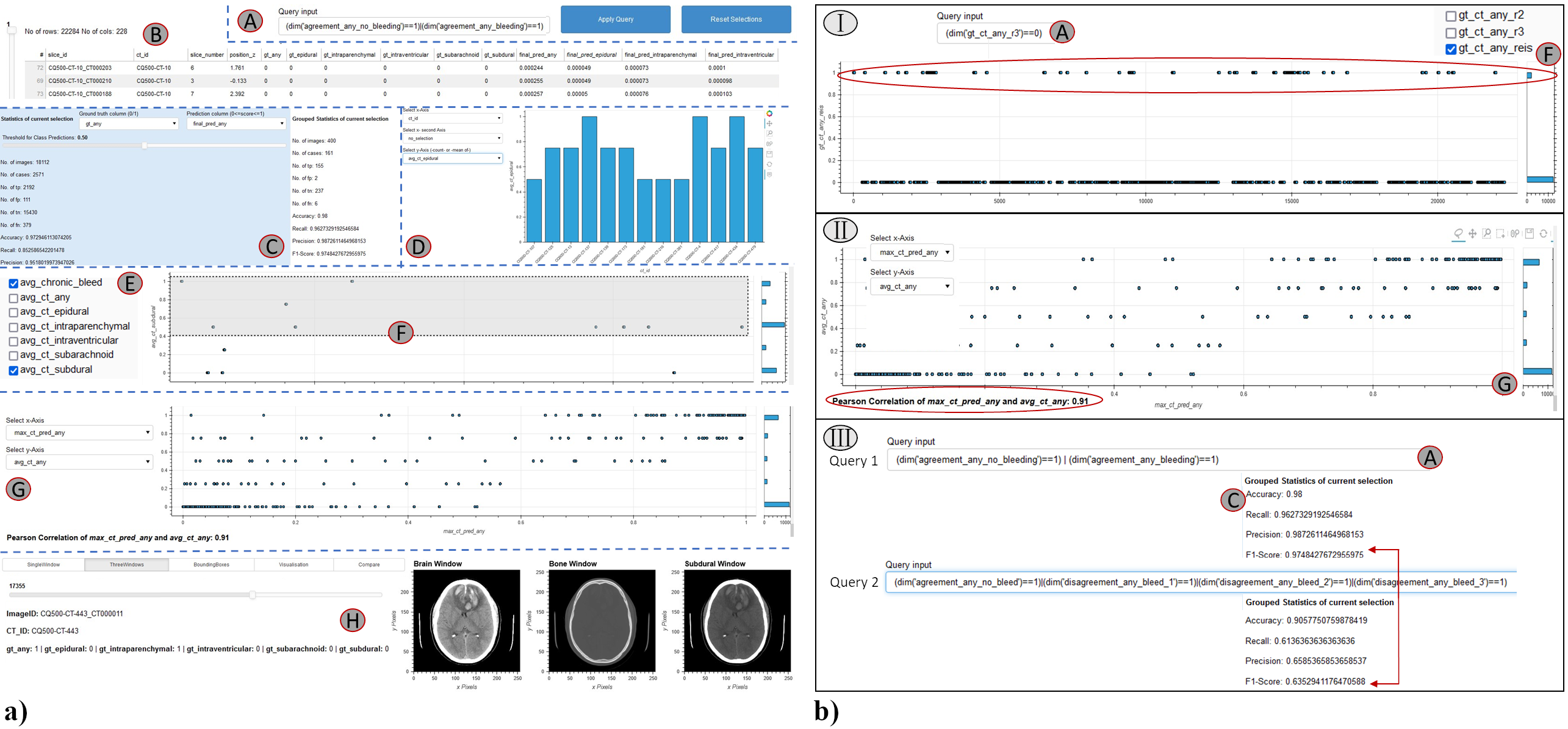}
    \caption{(a) Overview of the interactive functionalities in ScrutinAI applied to the medical use case. 
    (b) Detailed screenshots of the findings in analysis cycles I-III.
    (The individual widgets are shown in a compressed form to give an overall view.)}
    \label{fig:overview_scrutinAI}
\end{figure*}

Since labeling noise is a challenging and well-known problem in medical image analysis, we investigate the application of our model to the CQ500 data set in this regard.
Using ScrutinAI, we aim to expose dependencies and relationships between model performance and inter-observer variability in labeling. 
Therefore, we load the CQ500 data set, the annotations, and the precomputed model's predictions in \mbox{ScrutinAI}.
As visible in Figure~\ref{fig:overview_scrutinAI}a, all structured data is accessible over the meta data overview \circled{B}.
The unstructured image data can be displayed in use case specific views \circled{H}, enabling an analysis in different window settings or an inspection of annotated bounding boxes and local explanations generated with GradCAM++.
Following, we present the walk-throughs of four consecutive \textbf{analysis cycles} (I-IV) within ScrutinAI, where we gained six \textit{(i-iv)} findings. 

In \textbf{cycle I}, we want to compare the model's performance against the opinion of all four radiologists as ground truth.
For this task, we iteratively use the metric overview widget (as illustrated in Figure~\ref{fig:overview_scrutinAI}a \circled{C}) and interactively change which annotation column to use as ground truth. 
All other dependent performance metrics are re-computed automatically.
Based on this, we find that among radiologists, a varying number of hemorrhages and corresponding subtypes is detected.
The model shows the best performance compared to the annotations of the fourth radiologist (Acc: 92,4\%, F1-Score: 91\%, occasions class any: 217), and the worst performance in comparison to the third radiologist (Acc: 88\% and F1-Score 87,4\%, occasions class any: 193). 
This can also be explored visually in ScrutinAI using the widgets textual query \circled{A} and  scatter plot \circled{E}.
We select with the query all hemorrhages labeled as negative by one radiologist, and create three plots for the annotations of the other radiologists to visually compare patterns. 
Exemplary, one of these plots in Figure~\ref{fig:overview_scrutinAI}b - cycle I shows that several annotations of the fourth radiologist differ to the third expert and are labeled as true (value 1).
We gain the first two findings: \textbf{\textit{(i)}} the test data set shows a high inter-observer variability with respect to the general occurrence of hemorrhage as well as among different classes, and 
\textbf{\textit{(ii)}} the model's performance varies greatly depending on which radiologist it is compared to.

We deepen the analysis in \textbf{cycle II} and calculate for each CT how many radiologists agreed on the presence of a hemorrhage. 
In the drop-down menu of the pearson correlation plot \circled{G}, we choose the agreement of the radiologists on the incidence of a hemorrhage and the prediction of the model as independent categories and as result we get a positive correlation of 0.91 (see Figure~\ref{fig:overview_scrutinAI}b - cycle II). 
The plot visualizes, that in cases where radiologists agree on the occurrence of a hemorrhage, the prediction score of the model is higher, whereas the less radiologists agree on a hemorrhage, the lower the prediction score of the model.
We conclude as third finding \textbf{\textit{(iii)}} a correlation between the model's performance and the inter-observer variability.

Using again the textual query, we further explore this observation in \textbf{cycle III}. 
Based on the data selection, \mbox{ScrutinAI} provides the number of cases detected  by one, two, three or four radiologists, respectively, and let us easily compare how many of those potential cases are detected by our model. 
The results (see Table \ref{tab:agreement_on_label}) support finding \textit{\textbf{(iii)}}. 
Most cases are detected by all four radiologists and the detection rate of the model for those cases is the highest.
Detection overlap between model and radiologists decreases, if only three or two radiologists agree on the occurrence. 
For those cases in which only one radiologist detects a hemorrhage, the model classifies a hemorrhage even less frequently. 
\begin{table}[t]
    \centering
    \begin{tabular}{|c  c |c |c |c| c|}
    \hline
         &&Four &Three & Two &One\\
         \hline
          Cases & & 161 & 37 & 28 & 25\\
          \hline           
          Model & True & 155 & 25 & 11 & 3\\
            & False & 6 & 12 & 17 & 22\\
          \hline
          Detect. overlap  & & 97\% & 68\% & 61\%& 44\%\\
          \hline
    \end{tabular}
    \caption{Number of radiologists agreeing on a hemorrhage and model prediction for these potential cases.}
    \label{tab:agreement_on_label}
\end{table}
To compare the cases in which the radiologists agree with those in which they disagree, we query in \mbox{ScrutinAI} two data subsets (see Figure~\ref{fig:overview_scrutinAI}b - cycle III).
The first subset of data is selected by textual query 1, consisting of all cases on which all radiologists either agreed on ``no hemorrhage'' or ``hemorrhage''. 
The second subset is filtered analogously \mbox{(query 2)}, containing the data points where all radiologists agree on ``no hemorrhage'', plus, all cases for which only some of the radiologists (one, two or three) have detected a hemorrhage. 
We read off the model performance for both subsets by selecting the ground truth of the original annotations. 
Using the same procedure, we split the data according to the concordance of all radiologists within each class. 
We reveal as finding \textbf{\textit{(iv)}}, that accuracy and F1-score decrease in all classes for cases in which the radiologists do not agree on the label, while both metrics increase for all cases in which all radiologists agree on the label.

\begin{figure}[t]
\centering
\includegraphics[width=0.99\columnwidth]{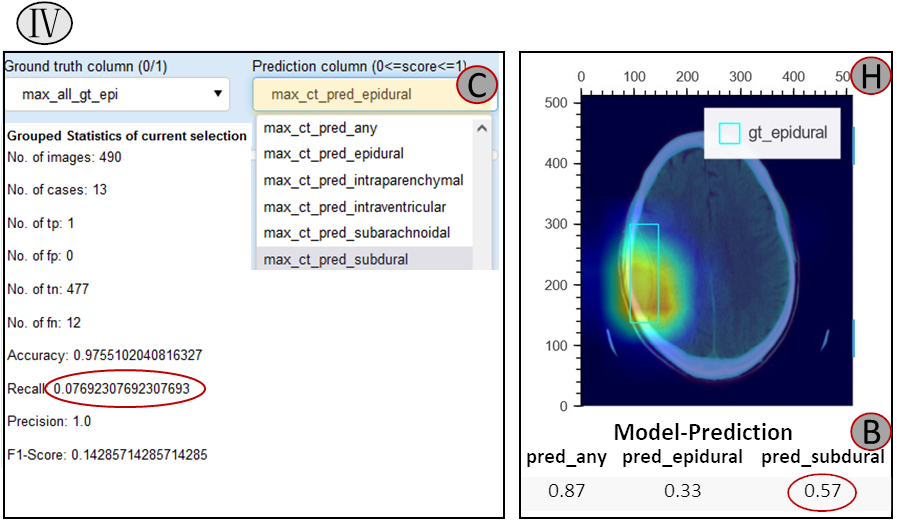} 
\caption{The findings of analysis cycle IV show in the performance overview and image view that the model incorrectly classifies epidural hemorrhages as subdural.} 
\label{fig:cycle4}
\end{figure}

As the performance of the model for epidural hemorrhages is extremely low, we scrutinize those cases in \textbf{cycle IV}. 
By examining the meta data overview, we observe an increased prediction score in the subdural class for many cases that have been annotated as epidural. 
As visible in Figure~\ref{fig:cycle4}, we explore in the image view \circled{H} an example case in which the model classifies a hemorrhage as subdural (see increased prediction score for subdural), while the ground truth of the slice and the bounding box are epidural.
Still, the local explanation with GradCAM++ shows that the model has detected the region within the bounding box but it has classified it as wrong subtype. 
As we are interested if the model systematically detects epidural hemorrhages as subdural, we select in the metric overview as visible in Figure \ref{fig:cycle4} \circled{C} as ground truth column ``epidural''.
Accordingly, we compare it firstly to the prediction column of epidural. 
As the grouped statistics show, there are 13 cases out of the 490 CT-Scans with an epidural hemorrhage, but only one case is detected by the model (true positive), while 12 epidural cases are not detected as epidural (false positive).
In a second step, we select in the drop down menu as prediction column ``subdural'', still comparing it to the epidural ground truth. 
We observe that now 11 out of the 13 epidural cases are detected and therefore classified as subdural.
The recall increases to 85\%.  
This result leads to finding \textbf{\textit{(v)}}, that the model did not learn to distinguish between the classes subdural and epidural.
A major reason for the bad performance might be due to the fact that the epidural cases were undersampled in the RSNA training data set. 
Still, we want to assess the radiologists' agreement on the label epidural, as distinguishing between both (spatially very close) regions requires a lot of expert domain knowledge. 
Filtering the epidural cases once more in \mbox{ScrutinAI}, we find that in 6 cases all radiologists, in 13 cases three, in 4 cases two, and in 9 cases only one radiologist agreed on an epidural hemorrhage.
We select with a textual query all cases labeled in the original ground truth as epidural but not as subdural aiming to exclude images with hemorrhages in both regions. 
We obtain only 7 cases. 
We visualize the radiologists' assessment for subdural in a scatter plot and detect that in 4 out of the 7 cases at least one radiologist still labeled the case as subdural. 
Even if the number of cases is not representative, it indicates as finding \textbf{\textit{(vi)}}  that a clear distinction between similar classes, as subdural and epidural, is even non-trivial for experienced radiologists and similarly leads to higher
inter-observer variation affecting model performance negatively.

%% file: 03_conclusion.tex
\section{Conclusion}
We have shown that ScrutinAI can be easily adapted to a new domain such as healthcare.
New data sets and models can be loaded and analysed with the existing structure and features. 
Based on the modular structure, the tool can be easily extended by use case specific features.
With the exemplary analysis of DNN detection of intracranial hemorrhages, we demonstrate that ScrutinAI can be used to interactively explore the dependencies between model performance and label noise in data sets. 
ScrutinAI's workflow and interactive functionalities were shown to efficiently support the analyst by means of VA principles. 
Using linked brushing, data could be easily filtered across the different widgets and data representations (structural and visual), enabling deep-dive analysis without the need to deal with individual scripts or tools to get the same functionality.

In summary, the analysis of the use case in \mbox{ScrutinAI} reveals the aleatoric uncertainty in the CQ500 data set, which interrelates with the inter-observer variability. 
Based on finding \textit{\textbf{(iii)}}, we assume that in the RSNA data set, used for the training of our model, a similar label noise consists. 
We face the challenge of discerning whether the label noise arises from hard to detect hemorrhages being detected only by individual experts or from radiologists misclassifying artifacts as hemorrhages. 
Moreover, analysis cycles III and IV revealed the difficulty for a clear and consistent distinction between specific and, in particular, similar regions among observers. 
The negative effect of label noise on model performance detected in finding \textbf{\textit{(iv)}} and \textbf{\textit{(v)}}, confirms that learning patterns for the model is more difficult due to inter-observer variability. 
The question raises, whether a model trained either on the annotations of only one expert, or on individual annotations of several consistent experts, would more easily learn a stable behaviour and perform better on hard to detect samples. 
To answer this question, we would need to compare our model to a model trained only on annotations of a single radiologist, which we leave open for future work.
For a deeper analysis of the actual correspondence of the detected hemorrhages, we plan to compare the location of the bounding boxes of \citeauthor{reis2020brain} to the rough location regions annotated in \citeauthor{chilamkurthy2018development} as well as to a location's approximation of the occurrences detected by the model and extracted from the heatmap generated through GradCAM++. 

%% file: 04_acknowledgement.tex
\section{Acknowledgement}
The development of this publication was supported by the Ministry of Economic
Ministry of Economic Affairs, Industry, Climate Action and Energy of the State of North Rhine-Westphalia as part of the flagship project ZERTIFIZIERTE KI. The authors would like to thank the consortium for the successful cooperation.